\documentclass[conference]{IEEEtran}
\IEEEoverridecommandlockouts
\usepackage{cite}
\usepackage{amsmath,amssymb,amsfonts}
\usepackage{textcomp}
\usepackage{xcolor}
\usepackage{color}
\usepackage{times}
\usepackage{epsfig}
\usepackage{graphicx}
\usepackage{amssymb}
\usepackage{verbatim}
\usepackage{graphicx}
\usepackage{subfig}
\usepackage{multirow}
\usepackage{array}
\usepackage{float}
\usepackage{multirow}
\usepackage{amsmath}
\usepackage{siunitx}
\usepackage[english]{babel}
\usepackage[utf8]{inputenc}
\usepackage{algorithm}
\usepackage[noend]{algpseudocode}
\usepackage{cite}
\usepackage{amsmath,amssymb,amsfonts}
\usepackage{textcomp}
\usepackage{xcolor}
\usepackage{color}
\usepackage{times}
\usepackage{epsfig}
\usepackage{graphicx}
\usepackage{amsmath}
\usepackage{amssymb}
\usepackage{verbatim}
\usepackage{graphicx}
\usepackage{subfig}
\usepackage{multirow}
\usepackage{array}
\usepackage{float}
\graphicspath{{images/}}
\makeatletter
\newcommand*{\rom}[1]{\expandafter\@slowromancap\romannumeral #1@}
\makeatother
\def\BibTeX{{\rm B\kern-.05em{\sc i\kern-.025em b}\kern-.08em
    T\kern-.1667em\lower.7ex\hbox{E}\kern-.125emX}}
\begin{document}

\title{Unsupervised Learning of Eye Gaze Representation\\ from the Web}
%{Unsupervised Eye Gaze Region Estimation}

\author{\IEEEauthorblockN{ Neeru Dubey \hspace{1cm} Shreya Ghosh \hspace{1cm} Abhinav Dhall }
\IEEEauthorblockA{
\textbf{L}earning \textbf{A}ffect and \textbf{S}emantic \textbf{I}mage analys\textbf{I}s (LASII) Group\\
\textit{Department of Computer Science and Engineering} \\
\textit{ Indian Institute of Technology Ropar}\\
 Ropar, India \\
\tt {\{neerudubey, shreya.ghosh, abhinav\} @iitrpr.ac.in}}

% \and

}

\maketitle
 
\begin{abstract}
Automatic eye gaze estimation has interested researchers for a while now. In this paper, we propose an unsupervised learning based method for estimating the eye gaze region. To train the proposed network ``Ize-Net'' in self-supervised manner, we collect a large `in the wild' dataset containing 1,54,251 images from the web. For the images in the database, we divide the gaze into three regions based on an automatic technique based on pupil-centers localization and then use a feature-based technique to determine the gaze region. The performance is evaluated on the Tablet Gaze and CAVE datasets by fine-tuning results of Ize-Net for the task of eye gaze estimation. The feature representation learned is also used to train traditional machine learning algorithms for eye gaze estimation. The results demonstrate that the proposed method learns a rich data representation, which can be efficiently fine-tuned for any eye gaze estimation dataset.   
\end{abstract}

% \begin{IEEEkeywords}
% component, formatting, style, styling, insert
% \end{IEEEkeywords}

\section{Introduction}
%\label{sec:intro}
The eye gaze estimation aims to determine the line-of-sight for the pupil. It provides information about human visual attention and cognitive process \cite{mason2004look}. It aids several applications such as human-computer interaction \cite{ghosh2018speech}, student engagement detection \cite{kaur2018prediction}, video games with basic human interaction \cite{barr2007video}, driver attention modelling \cite{fridman2016driver}, psychology research \cite{birmingham2009human} etc. 

Gaze estimation techniques can be broadly classified into two types: intrusive and non-intrusive. The intrusive technique requires contact with human skin or eyes. It includes usage of head-mounted devices, electrodes and  sceleral coils \cite{xia2007ir,robinson1963method,tsukada2011illumination}. These devices provide accurate gaze estimation but cause an unpleasant user experience. The non-intrusive technique does not require physical contact \cite{leo2014unsupervised}. Image processing based gaze estimation methods come under the non-intrusive category. These methods face a number of challenges, which include partial occlusion of the iris by the eyelid, illumination condition, head pose, specular reflection if the user wears glasses, etc; inability to use standard shape fitting for iris boundary detection; and effects like motion blur and over saturation of image \cite{leo2014unsupervised}. To deal with these challenges, most of the accurate gaze estimation methods have been performed under constrained environments like fixation of head pose, illumination conditions, camera angle, etc. Such methods require huge dump of high-resolution labelled images. Robust gaze estimation needs accurate pupil-center localization. Fast and accurate pupil-center localization is still a challenging task \cite{gou2017joint}, particularly for images with low resolution.

%   Some robust image processing \cite{villanueva2013hybrid} and learning-based \cite{campadelli2009precise} methods have also been proposed, but they require high-resolution images and the tedious task of image labeling for supervised machine learning. 

%We can directly use the gaze region information wherever applicable or we can also fine-tune the learned data representation for any specific dataset. We demonstrate the fine-tuning results for Tablet Gaze and CAVE datasets. These results demonstrate the effectiveness of the learned data representation.
With the success of supervised deep learning techniques, especially convolution neural networks, much progress has been witnessed in most of the problems in computer vision. This is primarily due to the availability of graphics processing unit (GPU) hardware and large-sized labelled databases. Furthermore, it has been noted that the labelling of complex vision task is a noisy and erroneous process. Hence, there is an interest in exploring deep learning based unsupervised techniques for computer vision tasks \cite{wang2015unsupervised, misra2016shuffle,zhou2017unsupervised, datta2018unsupervised}. 

In this paper, we propose an unsupervised (self-supervised) technique for learning a discriminative eye-gaze representation. The method is based on exploiting the domain knowledge generated by analyzing YouTube videos. The aim is to learn a feature representation for eye gaze, which can be easily used by itself or fine-tuned for complex eye gaze related tasks. The experimental results show the effectiveness of our technique in predicting the eye gaze as compared to supervised techniques.

%The domain knowledge is labeled signifying if a subject is looking towards his/her left, right, or center region.

\begin{figure*}[t]
\centering
\subfloat{\includegraphics[width = 1in,height=1in]{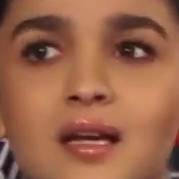}}\hspace{0.01in} 
\subfloat{\includegraphics[width = 1in,height=1in]{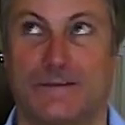}}\hspace{0.01in} 
\subfloat{\includegraphics[width = 1in,height=1in]{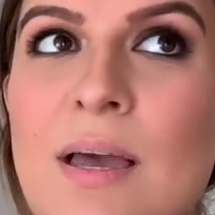}}\hspace{0.01in}
\subfloat{\includegraphics[width = 1in,height=1in]{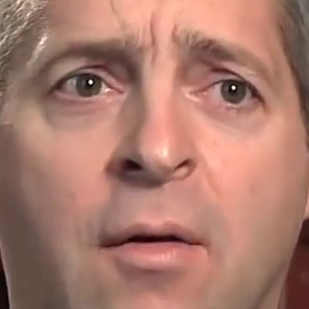}}\hspace{0.01in} 
\subfloat{\includegraphics[width = 1in,height=1in]{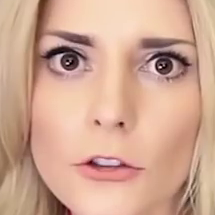}}\hspace{0.01in} 
\subfloat{\includegraphics[width = 1in,height=1in]{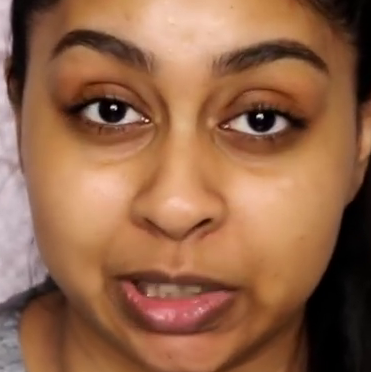}}\vspace{0.1in}
\subfloat{\includegraphics[width = 1in,height=1in]{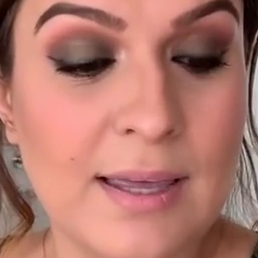}}\hspace{0.01in}
\subfloat{\includegraphics[width = 1in,height=1in]{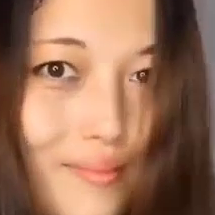}}\hspace{0.01in}
\subfloat{\includegraphics[width = 1in,height=1in]{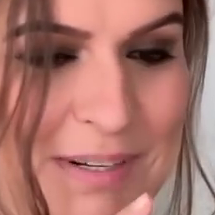}}\hspace{0.01in}
\subfloat{\includegraphics[width = 1in,height=1in]{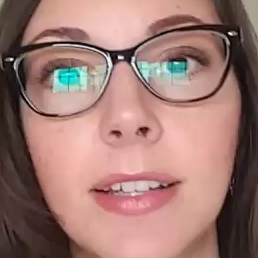}}\hspace{0.01in}
\subfloat{\includegraphics[width = 1in,height=1in]{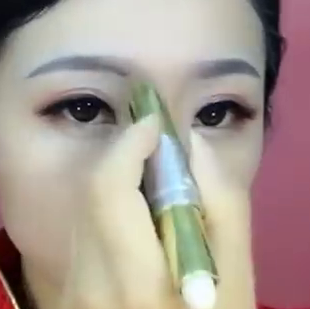}}\hspace{0.01in}
\subfloat{\includegraphics[width = 1in,height=1in]{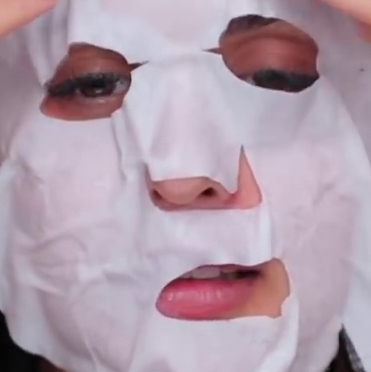}}\vspace{0.1in}
\subfloat{\includegraphics[width = 1in,height=1in]{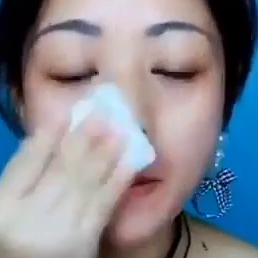}}\hspace{0.01in} 
\subfloat{\includegraphics[width = 1in,height=1in]{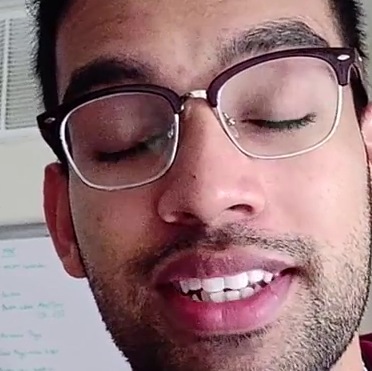}}\hspace{0.01in} 
\subfloat{\includegraphics[width = 1in,height=1in]{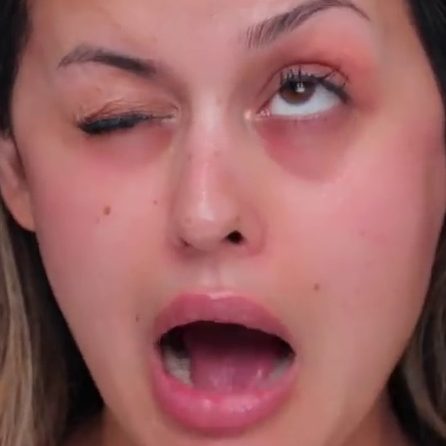}}\hspace{0.01in}
\subfloat{\includegraphics[width = 1in,height=1in]{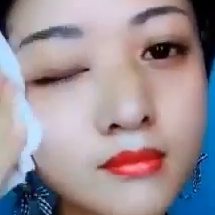}}\hspace{0.01in} 
\subfloat{\includegraphics[width = 1in, height=1in ]{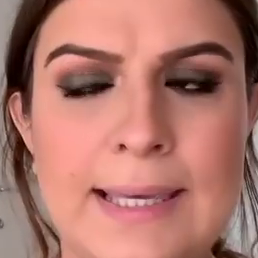}} \hspace{0.01in} 
\subfloat{\includegraphics[width = 1in,height=1in]{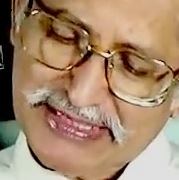}}\vspace{0.1in} 
\caption{\small Sample images from proposed dataset. Here, we can see that there is huge variation in illumination, facial attributes of subjects, specular reflection, occlusion, etc. First and second rows from top show images for which the gaze region is correctly estimated and third row shows images where gaze region is not correctly estimated. First row; first and second images are looking towards left region. First row; third and fourth images are looking towards right region. First row; fifth and sixth images are looking towards central region. Second row contains images of challenging scenarios like, occlusion and specular reflection; for which we get correct gaze region estimation. Last row contains images of scenarios where our method fails due to insufficient information for determining correct gaze region. (Image Source: YouTube creative commons)}
\label{fig:sample_images}
\end{figure*}

The main contributions of this paper are as follows:
\begin{enumerate}
\item Dataset (Figure \ref{fig:sample_images}) of 1,54,251 facial images of 100 different subjects from YouTube videos has been collected. These images are automatically labeled using the proposed method of eye gaze region estimation.
\item Propose a deep neural network, ``Ize-Net", which is trained on the proposed dataset. The result shows that unsupervised techniques can be used for learning rich representation for eye gaze.
\item Method to detect if the subject present in the input image, is looking towards his/her left, right, or center region. The gaze region estimation is calculated by utilizing the relative position of both (left and right) the pupils in the eye sockets.
\item Method to localize pupil-center using facial landmarks, OTSU thresholding \cite{otsu1979threshold} and  Circular Hough Transformation (CHT) \cite{pedersen2007circular}.
%\item Finally, demonstrate the effectiveness of learned features for any specific facial image dataset.
\end{enumerate}

The remainder of this paper is organized as follows: Section \ref{sec:Related Work} describes some of the related studies. Section \ref{sec:Proposed Method} presents the details of the proposed pupil-center localization and gaze estimation methods. In Section \ref{sec:Experiments}, we empirically study the performance of the proposed approach. Section \ref{sec:Conclusion and Future Work} contains the conclusion and future work.

\section{Related Work}
\label{sec:Related Work}
% \subsection{Gaze Tracking Techniques}
% In the recent past, various gaze tracking techniques have been proposed and applied in different fields \cite{hansen2010eye,rayner2001integrating,duchowski2002breadth}. The research in this field is mainly based on two areas: eye localization and eye gaze estimation \cite{hansen2010eye}.
% like virtual reality, human-attention modeling, human-computer interaction  \cite{hansen2010eye}, cognitive linguistics, marketing and advertisement \cite{rayner2001integrating}. The eye gaze information can be employed to be used by the disabled persons to control the computer effectively \cite{duchowski2002breadth}. 

% \subsubsection{Eye localization}
% Eye localization aims at finding the location of pupil-center in a given eye/face image.  In this paper, we propose a geometric feature-based method to locate pupil-centers in the images captured under real-world scenarios with various types of cameras.

The proposed method contains pupil-center localization and eye gaze estimation techniques. Accordingly, the literature survey demonstrates some of the relevant pupil-center localization and eye-gaze estimation methods. 

The most popular solutions presented for the task of pupil-center localization can be broadly classified into active and passive methods \cite{leo2014unsupervised}. 

The active pupil-center localization methods utilize dedicated devices to precisely locate the pupil-center such as infrared camera \cite{xia2007ir}, contact lenses \cite{robinson1963method} and  head-mounted device \cite{tsukada2011illumination}. These devices require pre-calibration phase to perform accurately. They are generally very expensive and cause uncomfortable user experience. 

The passive eye localization methods try to gather information from the supplied image/video-frame, regarding pupil-center. Valenti et al. \cite{valenti2012accurate}, have used isophotes to infer circular patterns and used machine learning for the prediction task. An open eye can be peculiarly defined by its shape and its components like iris and pupil contours. The structure of an open eye can be used to localize it in an image. Such methods can be broadly divided into voting-based methods \cite{kim1999vision,perez2003precise} and model fitting methods \cite{daugman2003importance,hansen2005eye}. Although these methods seem very intuitive, they do not provide good accuracy. Several machine learning based pupil-center localization methods have also been proposed. One such method was proposed by Campadelli et al. \cite{campadelli2009precise}, in which they used two Support Vector Machines (SVM) and trained them on properly selected Haar wavelet coefficients. In \cite{markuvs2014eye}, randomized regression trees were used. These supervised learning based methods require the tiresome process of data labeling. 

In this paper, we propose a method which overcomes the aforementioned limitations. It is a geometric feature-based pupil-center localization method, which gives accurate results for images captured under uncontrolled environment. In the past, various visible imaging based eye gaze tracking methods have been proposed which can be broadly classified among feature-based methods and appearance-based methods. 

Feature-based methods utilize some of the prior knowledge to detect the subject's pupil-centers from simple pertinent features based on shape, geometry, color and  symmetry. These features are then used to extract eye movement information. Morimoto et al. \cite{morimoto2000pupil} assumed a flat cornea surface and proposed a polynomial regression method for gaze estimation.  In \cite{zhu2002subpixel}, Zhu and Yang extracted intensity feature from an image and used a Sobel edge detector to find pupil-center. The gaze direction was determined via linear mapping function. The detected gaze direction was sensitive to the head pose, therefore, the users must stabilize their heads. In  \cite{torricelli2008neural}, Torricelli et al. performed the iris and corner detection to extract the geometric features, which were mapped to the screen coordinates by the general regression neural network. In \cite{valenti2012accurate}, Valenti et al. estimated the eye gaze by combining the information of eye location and head pose.

Appearance-based gaze tracking methods do not explicitly extract the features instead they utilize the whole image for eye gaze estimation. These methods, normally do not require the geometry information and calibration of cameras, since the gaze mapping is directly performed on the image content. These methods usually require a large number of images to train the estimator. To reduce the training cost, Lu et al. \cite{lu2014learning} proposed a decomposition scheme. It included the initial gaze estimation and the subsequent compensations for the gaze estimation to perform effectively using training samples.  Huang et al. \cite{huang2015tabletgaze} proposed an appearance based gaze estimation method in which the video captured from the tablet was processed using HoG features and Linear Discriminant Analysis (LDA). In \cite{lu2010novel}, an eye gaze tracking system was proposed, which extracted the texture features from the eye regions using the local pattern model. Then it fed the spatial coordinates into the Support Vector Regressor to obtain a gaze mapping function. Zhang et al. \cite{zhang2017mpiigaze} proposed GazeNet which was deep gaze estimation method. Williams et al. \cite{williams2006sparse} proposed a sparse and semi-supervised Gaussian process model to infer the gaze, which simplified the process of collecting training data.

In this paper, we propose an unsupervised method to detect whether a subject is looking towards his/her left, right or center region. We utilize the relative position of pupils in the eye sockets for judging the gaze region. This method allows us to predict the gaze region for a variety of images containing different textures, races, genders, specular attributes, illuminations and  camera qualities.

\section{Proposed Method}
\label{sec:Proposed Method}
This section explains the pipeline of proposed gaze region estimation method. At first, we localize pupil-centers. Then, utilize them to estimate the region of eye gaze using an intuitive approach which works well for images captured in the wild. Further, for learning eye gaze representation, we collect a large dataset of human faces. The domain knowledge based pupil-centers and facial points are used to create noisy labels representing gaze regions (\emph{left}, \emph{right}, or \emph{center}). A network is then trained for the mentioned task. Later we show the usefulness of the representation learned from the fully automatically generated noisy labels.

\subsection{Dataset Collection}
\label{sec:Dataset Collection}
In recent years, several gaze estimation datasets have been published. Most of the datasets contain very less variety of images in terms of head poses, illumination, number of images, collection duration per subject and  camera quality. To demonstrate that our proposed method is versatile, we collect a dataset containing 1,54,251 facial images belonging to 100 different subjects. The overall statistic of our dataset is shown in Table \ref{data_stat}. We download different types of videos from YouTube's creative common section. These videos are basically of the category where a single subject is seen on the screen at a time, like news reporting, makeup tutorials, speech videos, etc. We have considered every third frame of the collected videos for dataset creation. For the training purpose, the dataset has been split into training and validation sets with 70\% and 30\% uniform partitions over the subjects. The overview of our proposed dataset has been shown in Figure \ref{fig:sample_images}. In this figure, we can observe that our dataset contains a huge variety of images with varying illumination, occlusion, blurriness, color intensity, etc. Table \ref{subject_comparison} provides a comparison of the state-of-the-art gaze datasets with our proposed dataset.

\subsection{The Pupil-Center Localization}
\label{pupil_loc}
%We utilize the distinctive properties of the eye to determine the pupil-centers. 
Accurate pupil-center localization plays an important role in eye gaze estimation. We take face image as input and extract eyes from this image making use of the facial landmarks obtained by Dlib-ml library \cite{king2009dlib}. Further processing is performed on the extracted eye images. We localize pupil-center using two methods i.e. blob center detection and CHT; and take average of the pupil-centers obtained by both the methods to calculate the final pupil-center.

The steps of the proposed pupil-center localization method are explained below:
\begin{enumerate}
\item Extract eyes using facial landmarks information.
\item Apply OTSU thresholding on the extracted eyes to take advantage of unique contrast property of eye region while pupil circle detection.
\item Apply the method of blob center detection on extracted iris contours to calculate 'primary' pupil-centers.
\item Crop regions near these centers, to perform the center rectification task. The crop length is decided by applying equation (1). 
    \begin{equation}
        Crop\, length = \dfrac{Height\, of\, eye\, contour}{2} + offset
        %Crop\, length = (Height\, of\, eye\, contour/ 2)+ offset
    \end{equation}
\item Compute Adaptive thresholding and apply Canny edge detector \cite{canny1986computational} to make the iris region more prominent.
\item Apply CHT over the edged image to find secondary pupil-centers.
\item Compute average of primary and secondary pupil-centers to finalize the value for pupil-centers.
\end{enumerate}

\begin{figure*}[t]
\centering
\subfloat{\includegraphics[width = 1.7in,height=1in]{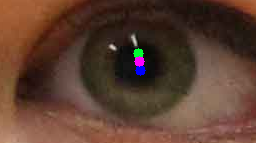}}\hspace{0.01in} 
\subfloat{\includegraphics[width = 1.7in,height=1in]{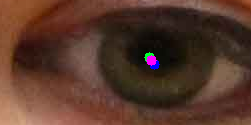}}\hspace{0.01in} 
\subfloat{\includegraphics[width = 1.7in,height=1in]{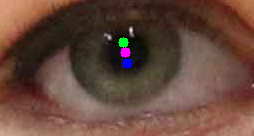}}\hspace{0.01in} 
\subfloat{\includegraphics[width = 1.7in,height=1in]{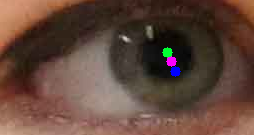}}
\caption{\small Results of pupil-center localization method. Green, blue and  pink colors represent the pupil-centers as mentioned in Section \ref{pupil_loc} (Image Source:\cite{smith2013gaze} best viewed in color).} 
\label{fig:pupil_eye}
\end{figure*}

\begin{figure}[htbp] 
    \centering
      \includegraphics[height=7.5cm,width=\linewidth]{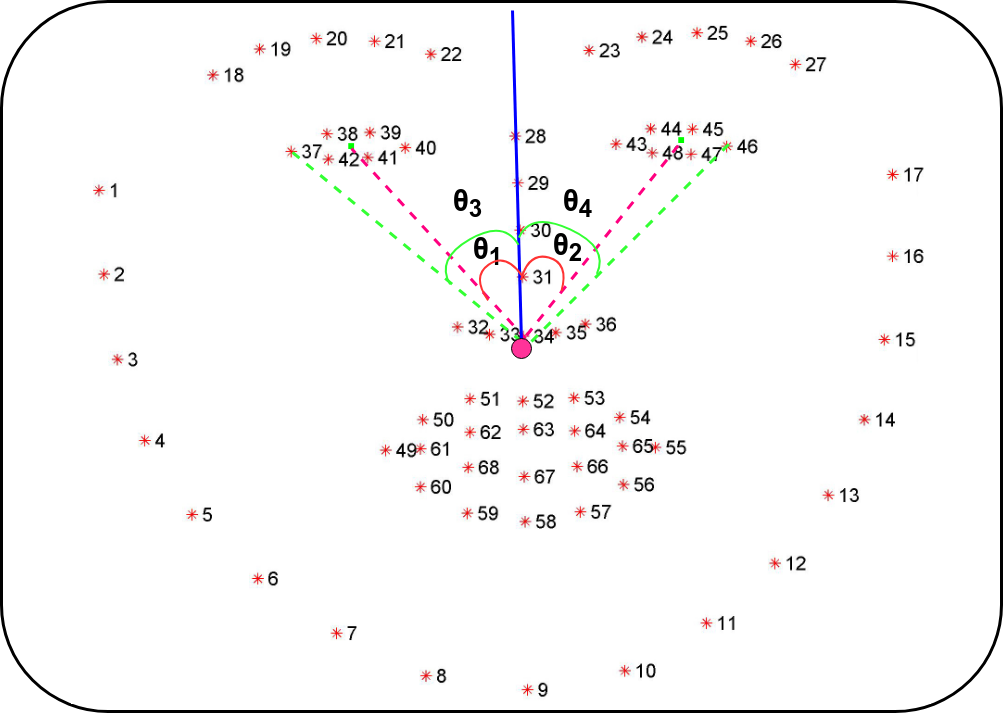}
      \caption{\small Facial points and corresponding four angles $\theta_1$, $\theta_2$, $\theta_3$ and  $\theta_4$ are depicted. Angles $\theta_1$ and $\theta_2$ are used to estimate eye gaze region and angles $\theta_3$ and $\theta_4$ are used for head pose estimation.}
      \label{fig:angle}
      \vspace{-3mm}
\end{figure}

Empirically, we noticed that the pupil-center localization accuracy is increased by taking an average of pupil-centers calculated by the above two methods.
Few sample results of pupil-center localization have been shown in Figure \ref{fig:pupil_eye}. The blue, green and pink dots represent the pupil-center obtained by our primary method, secondary method and their average, respectively.

\subsection{Heuristic for Eye Gaze Region Estimation}
\label{heuristic}
The Pupil-center is the most decisive feature of the face to determine gaze direction. Eyeballs move in the eye sockets to change the direction of the gaze. By using the relative position of both the pupil-centers, we can determine the region in which the subject is looking. When a subject looks towards his/her left, both the eyes' iris shift towards left. To utilize this characteristic, we compare the angles which are formed when we join left pupil-center with nose and nose with vertical; with the angle which is formed when we join right pupil-center with nose and nose with vertical. These angles are demonstrated in Figure \ref{fig:angle} as angles $\theta_1$ and $\theta_2$. For a subject to look towards his/her left region, the left eye angle $\theta_1$ has to be bigger than the right eye angle $\theta_2$. This intuitive heuristic is used to detect the region (left, right, or center) in which the subject is looking. The proposed method is immune to head movements within the range of \ang{-10} to \ang{10}.

The eye corners remain fixed with the eye movement. We utilize the eye corner points, given by Dlib-ml library, to determine the head pose direction, in the same way as we determine the eye gaze region. The angles used to determine the head pose direction are demonstrated in Figure \ref{fig:angle} as angles $\theta_3$ and $\theta_4$. 

\begin{table}[b]
\centering
\scalebox{1.2}{
\begin{tabular}{|c|c|c|c|c|}
\hline
\textbf{Dataset} & \textbf{Center} & \textbf{Left} & \textbf{Right} & \textbf{Total} \\ \hline
Train set        & 32,450           & 38,230         & 37,338          & 1,08,018         \\ \hline
Validation set   & 14,008           & 16,584         & 15,641          & 46,233          \\ \hline
Total            & 46,458           & 54,814         & 52,979          & 1,54,251         \\ \hline
\end{tabular}}
\caption{\small Proposed dataset's set vise distribution.}
\label{data_stat}
\end{table}

\begin{figure*}[t]
    \centering
      \includegraphics[height=5cm,width=\linewidth]{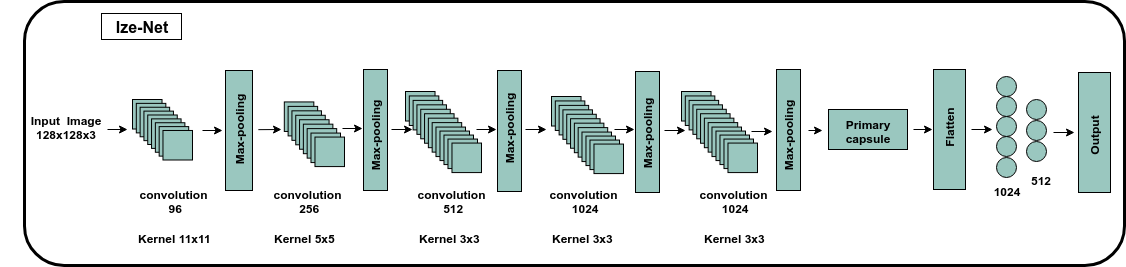}
      \caption{\small Overview of the proposed pipeline.} 
      \label{fig:pipeline}
\end{figure*}

\begin{table*}[t]
\centering
\scalebox{1.2}{
\begin{tabular}{|c|c|c|c|c|c|c|c|c|}
\hline
\textbf{Dataset}  & \textbf{\begin{tabular}[c]{@{}c@{}}Gi4E\\ \cite{villanueva2013hybrid}\end{tabular}} & \textbf{\begin{tabular}[c]{@{}c@{}} \cite{weidenbacher2007comprehensive}\end{tabular}} & \textbf{\begin{tabular}[c]{@{}c@{}}CAVE\\ \cite{smith2013gaze}\end{tabular}} & \textbf{\begin{tabular}[c]{@{}c@{}}OMEG\\ \cite{he2015omeg}\end{tabular}} &\textbf{\begin{tabular}[c]{@{}c@{}}MPII Gaze\\ \cite{zhang15_cvpr}\end{tabular}}&\textbf{\begin{tabular}[c]{@{}c@{}}Tablet Gaze\\ \cite{huang2015tabletgaze}\end{tabular}}&\textbf{\begin{tabular}[c]{@{}c@{}}GazeCapture\\ \cite{cvpr2016_gazecapture}\end{tabular}}& \textbf{Ours} \\ \hline
Subjects                                                                          & 103                                                         & 20                                                                        & 56                                                                                                       & 50  & 15 & 41 & 1450               & \textbf{100}           \\ \hline
\begin{tabular}[c]{@{}c@{}} Total  Images\end{tabular}                                                                              & 1,339                                                         & 2,220                                                                        & 5,880                              & 44,827             & 2,13,659  & 1,00,000  &   2,445,504               & \textbf{1,54,251}           \\ \hline
\end{tabular}}
\caption{\small Eye gaze datasets statistics.}
\label{subject_comparison}
\end{table*}

\subsection{Proposed Network Architecture}
Deep Neural Network (DNN) is well known to perform exceptionally well at handling visual recognition tasks. DNN predicts a face by detecting a bunch of randomly assembled face parts. Therefore, it is not suitable to be used when we utilize the relative position of face parts for classification purpose. To overcome this problem, the Capsule Network \cite{sabour2017dynamic} was proposed. Capsule Network constraints the relative position of face parts. The proposed method takes the face symmetry into consideration while detecting the eye gaze region. To combine the advantages of both the networks, we propose ``Ize-Net" network. The architecture of the proposed network is shown in Figure \ref{fig:pipeline}. This network is trained using images of size $128\times128\times3$. We have taken entire face as input instead of only eyes. According to \cite{zhang2017s}, gaze can be more accurately predicted when the entire face is considered. Our proposed network contains five Convolution layers. Each Convolution is followed by Batch Normalization and Max-Pooling. For Batch Normalization, we use 'ReLU' as the activation function. For Max-Pooling kernel of size ($2\times2$) was used. The stride of ($1\times1$) is considered for each layer. After the Convolution layers, we append Primary Capsule, whose job is to take the features learned by Convolution layers and produce combinations of the features to consider face symmetry into account. The output of the Primary Capsule is flattened and fed to Fully-Connected (FC) layers of dimension 1024 and 512. In the end, we apply `Softmax' activation to produce the final output. 
\begin{table}[t]
\centering
\scalebox{1.2}{
\begin{tabular}{|l|c|c|}
\hline
\textbf{Method/ Network} & \textbf{CAVE} & \textbf{\begin{tabular}[c]{@{}c@{}}Our\\ Dataset\end{tabular}} \\ \hline
Eye Gaze heuristic       & 60.37\%              & N/A                                                            \\ \hline
Alexnet                  & N/A              & 88.22\%                                                         \\ \hline
VGG-Face                  & N/A              & 84.30\%                                                         \\ \hline
Ize-Net                  & 82.80\%        & 91.50\%                                                        \\ \hline
\end{tabular}}
\caption{\small Validation of our proposed heuristic and Ize-Net network for CAVE dataset and proposed dataset.}
\label{result}
\end{table}

\begin{table*}[]
\setlength\extrarowheight{5pt}
\fontsize{11}{9}\selectfont
\centering
\scalebox{0.73}{
\begin{tabular}{|c|c|c|c|c|c|c|c|c|c|c|}
\hline
\begin{tabular}[c]{@{}c@{}} \textbf{Methods}\\ \textbf{Tablet Gaze} \end{tabular}& \begin{tabular}[c]{@{}c@{}}\textbf{Raw pixels}\\ \cite{huang2015tabletgaze}\end{tabular} & \begin{tabular}[c]{@{}c@{}}\textbf{LoG}\\ \cite{huang2015tabletgaze} \end{tabular} & \begin{tabular}[c]{@{}c@{}}\textbf{LBP}\\ \cite{huang2015tabletgaze} \end{tabular} & \begin{tabular}[c]{@{}c@{}}\textbf{HoG}\\ \cite{huang2015tabletgaze} \end{tabular} & \begin{tabular}[c]{@{}c@{}}\textbf{mHoG}\\ \cite{huang2015tabletgaze} \end{tabular} & \begin{tabular}[c]{@{}c@{}}\textbf{Ours}\\ \textbf{(full network}\\ \textbf{ fine tuning})\end{tabular} & \begin{tabular}[c]{@{}c@{}}\textbf{Ours} \\ \textbf{(last} \\ \textbf{12 layers} \\\textbf{fine-tuning)} \end{tabular} & \begin{tabular}[c]{@{}c@{}}\textbf{Ours} \\ \textbf{(last} \\ \textbf{8 layers} \\\textbf{fine-tuning)}\end{tabular} & \begin{tabular}[c]{@{}c@{}}\textbf{FC (34)} \\ \textbf{+ SVR} \end{tabular} & \begin{tabular}[c]{@{}c@{}}\textbf{FC (31)} \\ \textbf{+ SVR} \end{tabular} \\ \hline
k-NN               & 9.26       & 6.45 & 6.29 & 3.73 & 3.69 & \multirow{4}{*}{2.36}                                                                     & \multirow{4}{*}{3.31}                                                                                    & \multirow{4}{*}{3.26}                                                                                          & \multirow{4}{*}{2.42}                                            & \multirow{4}{*}{2.48}                                           \\ \cline{1-6}
RF                 & 7.2        & 4.76 & 4.99 & 3.29 & 3.17 &                                                                                            &                                                                                                          &                                                                                                            &                                                                   &                                                                   \\ \cline{1-6}
GPR                & 7.38       & 6.04 & 5.83 & 4.07 & 4.11 &                                                                                            &                                                                                                          &                                                                                                            &                                                                   &                                                                   \\ \cline{1-6}
SVR                & -          & -    & -    & -    & 4.07 &                                                                                            &                                                                                                          &                                                                                                            &                                                                   &                                                                   \\ \hline
\end{tabular}}
\caption{\small Results on Tablet Gaze with comparison to baselines \cite{smith2013gaze}. Effectiveness of learnt features in Ize-Net is demonstrated by the fine tuning the network and by training a SVR over various FC layer features. }
\label{tg_compare}
\end{table*}

\begin{table*}[!htbp]
\centering
\scalebox{1.2}{
\begin{tabular}{|c|c|c|c|c|c|}
\hline
\textbf{Calibration}                                                                          & \multirow{2}{*}{\textbf{Method}} & \multicolumn{2}{c|}{\textbf{$\ang{0}$ yaw angle}} & \multicolumn{2}{c|}{\textbf{Full Dataset}} \\ \cline{1-1} \cline{3-6} 
\multirow{4}{*}{\begin{tabular}[c]{@{}c@{}}5 point system\\ (cross arrangement)\end{tabular}} &                                  & \textbf{X}           & \textbf{Y}          & \textbf{X}           & \textbf{Y}          \\ \cline{2-6} 
                                                                                              & Skodras et al. \cite{skodras2015visual}           & $2.65 \pm 3.96$          & $4.02 \pm 5.82        $ & N/A                  & N/A                 \\ \cline{2-6} 
                                                                                              & Jyoti et al. \cite{jyoti2018automatic}                         & $2.03 \pm 3.01 $         & $3.47 \pm 3.99 $        & N/A                  & N/A                 \\ \cline{2-6} 
                                                                                              & Ours                             & $2.94\pm 2.16 $         & $2.74\pm1.92 $          & $1.67\pm1.19$            & $1.74\pm1.57          $ \\ \hline
\end{tabular}}
\caption{\small Results on the CAVE dataset using the angular deviation, calculated as mean error $\pm$ standard deviation (in degree).}
\label{cave_compare}
\end{table*}
\subsection{Dataset Specific Fine Tuning}
The Ize-Net network is trained on the proposed dataset for the task of gaze region estimation. The learned data representation can be fine-tuned over any specific dataset for determining the exact gaze location. In the experiments section, we demonstrate the various level of fine tuning results for Tablet gaze and CAVE datasets. The high accuracy of the experimental results demonstrate that the proposed method learns a rich data representation. The learned data representation can be directly used for gaze region estimation or it can also be fine tuned for exact gaze estimation task over a specific dataset.

\section{Experiments}
\label{sec:Experiments}
For experimental purpose, we use the Keras deep learning library with the tensorflow backend. The code, dataset and model is available online\footnote{https://github.com/Neerudubey/Unsupervised-Eye-gaze-estimation}.

\subsection{Validation of Pupil Localization}
The pupil-center detection is performed using OTSU thresholding with blob center detection and CHT. To perform CHT, we crop the image around the pupil-center which we detect using OTSU thresholding and blob-center. We use offset of 5 pixels to crop the image. We validate the proposed pupil-center localization method (Section \ref{pupil_loc}) on BioID dataset \cite{jesorsky2001robust}. BioID is a publicly available dataset which contains 1,521 frontal face images of 23 subjects. The evaluation protocol is mentioned in equation \ref{eqn:error}, is same as the one used in \cite{jesorsky2001robust}. 
\begin{equation}
\label{eqn:error} 
e=\dfrac{max(d_l-d_r)}{\lVert C_l-C_r \rVert}
\end{equation}
where, $e$ is the error term, $d\textsubscript{l}$ and $d\textsubscript{r}$ are the Euclidean distances between the localized pupil-centers and the ground truth ones; $C\textsubscript{l}$ and $C\textsubscript{r}$ are left and right pupil-centers respectively in the ground truth.

We neglect some of the images, where Dlib-ml failed to detect the face or any of the eye contours. Table \ref{pupil_val} shows the comparison of the proposed method with some of the state-of-the-art methods. This table shows that our method is absolutely accurate in \begin{math} e \leq 0.10 \end{math} and \begin{math} e \leq 0.25 \end{math} cases, but it does not perform well enough when \begin{math} e \leq 0.05 \end{math}. The reason behind this is the inaccurate circle detection by CHT which propagates the error while averaging primary and secondary pupil-centers (Section \ref{pupil_loc}).

\begin{table}[b]
\centering
\scalebox{1.15}{
\begin{tabular}{|l|c|c|c|}
\hline
\multirow{2}{*}{\textbf{Methods}} & \multicolumn{3}{c|}{\textbf{Accuracy (\%)}}               \\ \cline{2-4} 
                                  & \textbf{\begin{math} e \leq 0.05 \end{math}} & \textbf{\begin{math} e \leq 0.10 \end{math}} & \textbf{\begin{math} e \leq 0.25 \end{math}} \\ \hline
\textbf{Ours}               & \textbf{56.97}    & \textbf{100.00}     & \textbf{100.00}      \\ \hline
Poulopoulos et al. \cite{poulopoulos2017new}                      & 87.10              & 98.00               & 100.00               \\ \hline
Leo et al. \cite{leo2014unsupervised}                             & 80.70              & 87.30             & 94.00                \\ \hline
Campadelli et al. \cite{campadelli2006precise}                       & 62.00                & 85.20             & 96.10              \\ \hline
Cristinacce et al. \cite{cristinacce2004multi}                     & 57.00                & 96.00               & 97.10              \\ \hline
Asadifard et al. \cite{asadifard2010automatic}                        & 47.00                & 86.00               & 96.00                \\ \hline
\end{tabular}}
\caption{\small Comparison of proposed pupil-center localization method with other state-of-the-art methods.}
\label{pupil_val}
\end{table}

\subsection{Validation of Eye Gaze Region Estimation}
The efficiency of the proposed eye gaze region estimation heuristic is validated on CAVE dataset \cite{smith2013gaze}. For this purpose, we map the given angular labels of CAVE dataset images into left, right and  center gaze regions based on the sign (positive and negative) of the gaze point mentioned. The validation results are shown in Table \ref{result}. After the heuristic evaluation, we also evaluate the performance of Alexnet \cite{krizhevsky2012imagenet} and VGG-Face \cite{parkhi2015deep} networks on the collected dataset. It gives 88.22\% validation accuracy for Alexnet and 84.30\% validation accuracy for VGG-face. For training both the networks, we use Stochastic Gradient Descent (SGD) optimizer with categorical cross-entropy as loss function. The learning rate and momentum are assigned 0.01 and 0.9 values respectively. 

\subsection{Performance of Ize-Net Network}
For training the proposed Ize-Net network, we initialize the network weights with `glorot normal' distribution. We use SGD optimizer with learning rate 0.001 with the decay of $1 \times e ^{-6}$ per epoch. We use categorical cross-entropy as loss function to train the proposed network. As mentioned in TABLE \ref{result}, it gives 91.50\% accuracy on the validation data of the proposed dataset. The proposed network outperforms the efficiency of AlexNet and VGG-face networks. The primary reason behind the better performance of Ize-Net is the presence of the primary capsule. This enables the network to consider the geometry of face into account during gaze region prediction. The consideration of face geometry is in accordance with the proposed heuristic used to label the images of the collected dataset. We validate the performance of the proposed network on CAVE dataset. The angular labels of CAVE dataset images have been mapped into three gaze regions. Post categorizing the images into their corresponding gaze regions we fine tune the Ize-Net for entire CAVE dataset to cross-check the performance of this network. We fine-tune our network for 10 epochs with 0.0001 learning rate \cite{smith2013gaze}. As mentioned in TABLE \ref{result}, our network gives 82.80\% five-fold cross-validation accuracy on CAVE dataset.

\subsection{Fine Tuning Results on Tablet Gaze and CAVE Datasets}
To fine tune the base model for prospective datasets, we add two Fully-Connected (FC) layers at the end of the proposed Ize-Net network. We fine-tune the network on Tablet Gaze and CAVE datasets. The two FC layers added in the base network are each of dimension 256 for both the datasets. The fine tuning results for Tablet Gaze are shown in TABLE \ref{tg_compare} and those for CAVE are shown in TABLE \ref{cave_compare}. As depicted in TABLE \ref{tg_compare} and \ref{cave_compare}, we  demonstrate the results for different levels of fine tuning. Last 8 FC layers, last 12 FC layers and complete network are fine-tuned one-by-one for the empirical analysis of results. For fine-tuning the proposed network for Tablet Gaze dataset, we used a learning rate of 0.0001 with 10 epochs and for CAVE dataset, we used a learning rate of 0.0001 with 15 epochs. For both the datasets, the fine tuning is done with mean square error loss function. The experimental results demonstrate that the proposed method outperforms the state-of-the-art gaze prediction for both Tablet Gaze and CAVE datasets. For experiments, we try our best to follow the protocols discussed in \cite{skodras2015visual} and \cite{huang2015tabletgaze}. However, there can be a few differences in frame extraction and selection. To demonstrate that the network learned efficient features, we trained a Support Vector Regressor (SVR) over the features learned in 31\textsuperscript{st} FC layer and 34\textsuperscript{th} FC layer for Tablet Gaze dataset. As depicted in TABLE \ref{tg_compare}, the low gaze prediction errors of SVR confirms that the learned features are highly efficient.

% \begin{table}[]
% \centering
% \begin{tabular}{|c|c|c|c|}
% \hline
% \textbf{Calibration}                                                                          & \multirow{2}{*}{\textbf{Method}} & \multicolumn{2}{c|}{\textbf{\ang{0} yaw angle}} \\ \cline{1-1} \cline{3-4} 
% \multirow{4}{*}{\begin{tabular}[c]{@{}c@{}}5 point system\\ (cross arrangement)\end{tabular}} &                                  & \textbf{X}           & \textbf{Y}          \\ \cline{2-4} 
%                                                                                               & Skodras et al. \cite{skodras2015visual}             & 2.65 \pm 3.96          & 4.02 \pm 5.82         \\ \cline{2-4} 
%                                                                                               & Jyoti et al. \cite{jyoti2018automatic}                           & 2.03 \pm 3.01          & 3.47 \pm 3.99         \\ \cline{2-4} 
%                                                                                               & Ours                             & 2.94 \pm 2.16                     &    2.74\pm 1.92                 \\ \hline
% \end{tabular}
% \caption{Result on the CAVE dataset using the angular deviation, calculated as mean error $\pm$ standard deviation (in degree
% ).}
% \label{cave_compare}
% \end{table}

\section{Conclusion and Future Work}
\label{sec:Conclusion and Future Work}
In this paper, we propose a method which learns a rich eye gaze representation by using unsupervised learning technique. Using the relative position of pupil-centers in left and right eye, the images are labeled based on gaze region i.e. left, right, or center. To demonstrate the robustness of the proposed method, we collect a large dataset of facial image. We also propose Ize-Net network, which is trained on the collected dataset. The weights of this trained model can be used for any facial image to detect the region of gaze. Machine learning methods can be used on the learned gaze region  representation to calculate eye gaze. Experimental results confirm the efficiency of the proposed method.

The proposed gaze estimation method can be vastly used for many human-computer interaction based applications without prior need of troublesome data labelling task.

% We demonstrate that the proposed method learns a highly efficient data representation by fine-tuning the network on Tablet Gaze and CAVE datasets. We also showcase the efficiency of learned features of FC layers of Ize-Net network by training SVR. 

Currently, our method is robust to the head pose movement within \ang{-10} to \ang{10}. In the future, we plan to utilize the head pose information completely while estimating the gaze region. We also plan to perform the real-time pupil-center localization and gaze region estimation for a video-based dataset.

\section*{Acknowledgement}
We gratefully acknowledge the support of NVIDIA Corporation with the donation of the Titan Xp GPU used for this research. 
% We thank our LASII lab members, who provided insight and expertise that greatly assisted this research.

\bibliographystyle{IEEEtran}  \small
\bibliography{eye_gaze}

\end{document}